\documentclass[letterpaper, 10 pt, conference]{ieeeconf}  

\IEEEoverridecommandlockouts                              

\usepackage{amsmath}       
\usepackage{amssymb}      
\usepackage{mathtools}
\usepackage{etoolbox}    
\usepackage{algorithm}
\usepackage{algorithmicx}
\usepackage{graphicx}
\usepackage{algpseudocode}
\usepackage{float}
\usepackage{booktabs} 
\usepackage{siunitx}
\usepackage{xcolor}
\usepackage{subcaption}

\usepackage{tabularx}
\usepackage{cite}
\usepackage{bm}

\title{\LARGE \bf
Grasp, Handover, Rotate: Bimanual Object Reorientation via Compositional Diffusion and Energy-Based Optimization
}

\author{
Wun Lam Yeung$^{1,*}$, 
Wenjun Liu$^{1,*}$, 
Yui Cheung Yu$^{1,*}$, 
Zhengyan Lambo Qin$^{1,*}$, 
Qijin She$^{1,2,\dagger}$, \\
Heng Li$^{1}$, 
Ziqi Wang$^{1}$, 
Ping Tan$^{1,2,\ddagger}$ \\[2mm]
$^1$The Hong Kong University of Science and Technology \\
$^2$Shenzhen Loop Area Institute
}

\begin{document}

\IEEEaftertitletext{%
  \begin{center}
    \includegraphics[width=\textwidth]{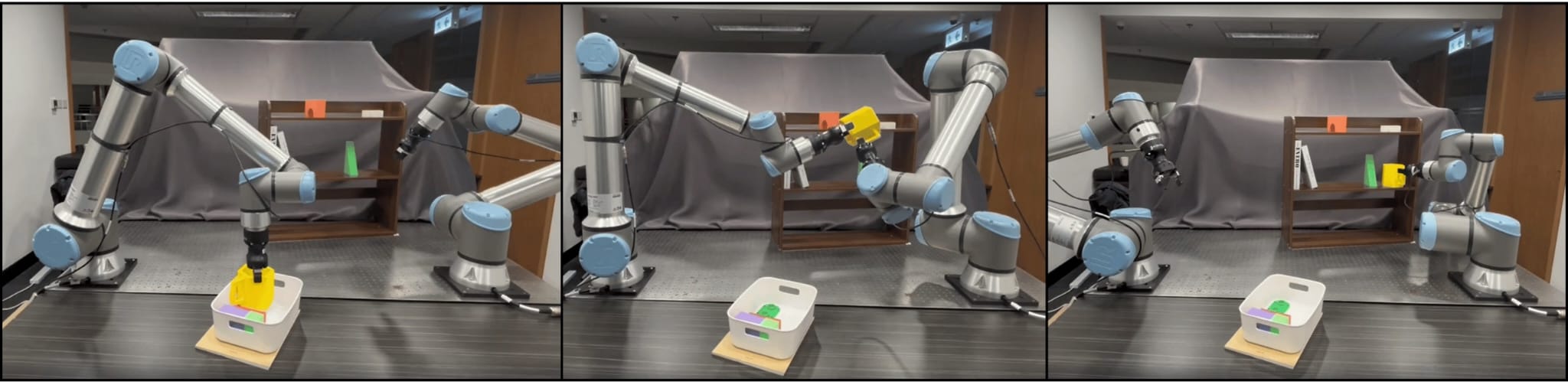}
    \captionof{figure}{Bimanual reorientation of a yellow cup using BiCompoDiff: the left arm picks and hands over the object, enabling the right arm to regrasp and place it in a reoriented pose with smooth, collision-free coordination.}
    \label{fig:teaser}
  \end{center}
  \vspace{0.5em}
}

\maketitle
\thispagestyle{empty}
\pagestyle{empty}

{\renewcommand{\thefootnote}{}
\footnotetext{$^*$ Co-first authors \quad $^{\dagger}$ Project lead \quad $^\ddagger$ Corresponding author: \texttt{pingtan@ust.hk}}
}

\begin{abstract}

Bimanual object reorientation—picking an object, handing it over between two arms, and placing it in a desired target pose—is valuable when direct placement from the initial grasp is infeasible due to collisions, kinematic constraints, or poor final orientation. However, achieving this under multiple competing objectives remains challenging. We introduce BiCompoDiff, a compositional diffusion and energy-based framework that jointly optimizes grasp selection, handover, regrasp, and motion planning under multiple constraints. By combining a pretrained grasp diffusion model with bimanual planning energy-based models (EBMs), our method injects gradient guidance during reverse diffusion to enforce collision avoidance, trajectory smoothness (via differentiable inverse kinematics), handover feasibility, and regrasp safety. Annealed MCMC sampling further refines grasp poses over the composite energy landscape. Experiments across diverse simulated household reorientation tasks demonstrate that BiCompoDiff achieves over 20\% higher success rates and up to 37\% smoother trajectories (measured by joint displacement) compared to strong sampling-based baselines. Real-world validation confirms effective sim-to-real transfer and robust performance on challenging scenes.
\end{abstract}

\section{INTRODUCTION}
\begin{figure}[!ht]
    \centering
    \includegraphics[width=0.9\linewidth]{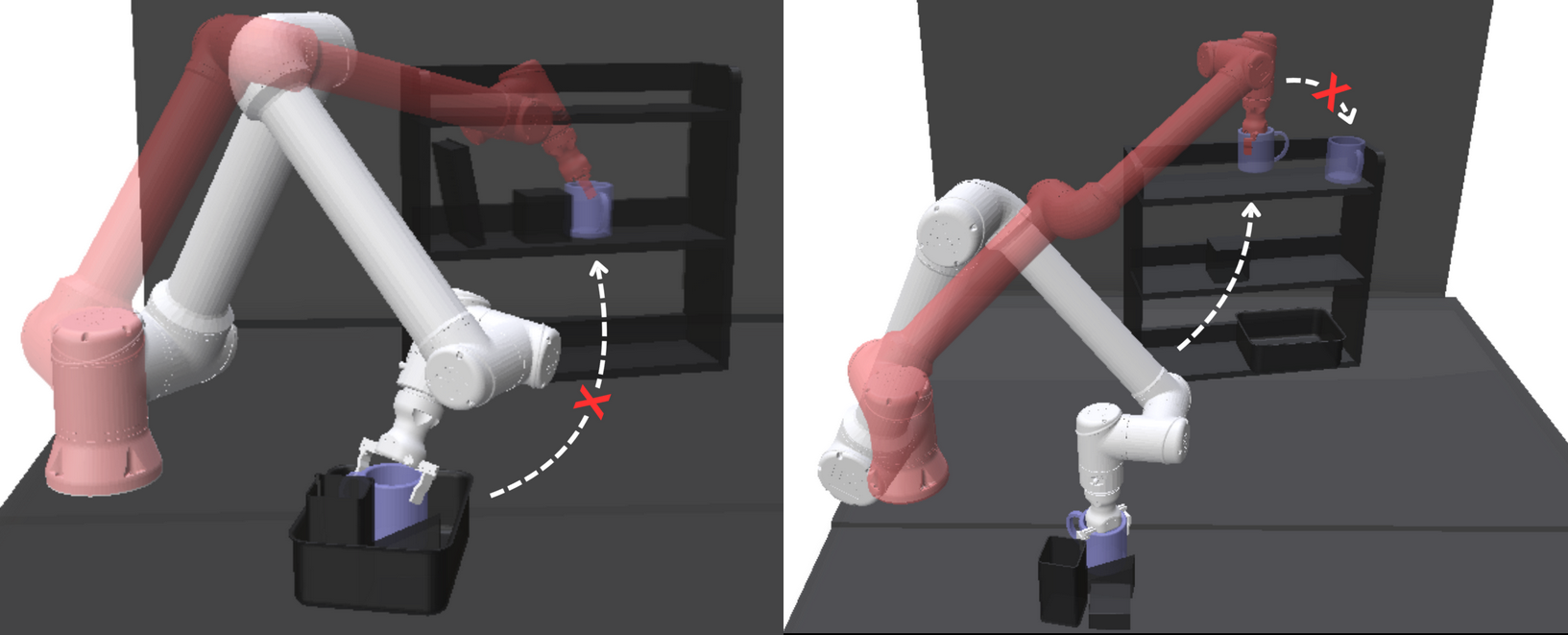}
    \caption{Scenarios requiring handover in dual-arm manipulation. 
    \textbf{Left:} No compatible collision-free grasp directions exist for both pick and place with a single arm. 
    \textbf{Right:} A single arm cannot find kinematically feasible grasp solutions for both the pick and place poses simultaneously.}
\label{fig:handover_needed}
\end{figure}

Rearranging objects into specific poses is a fundamental task in robotics, critical for both domestic tasks and industrial assembly. However, due to potential collisions and the inherent constraints of a robot's workspace, directly moving from the initial grasp pose to the target placement pose is often infeasible; a grasp that is convenient for initial acquisition is rarely suitable for final placement (as shown in Figure \ref{fig:handover_needed}). In such cases, object reorientation becomes an essential intermediate manipulation step: changing the object's pose to satisfy goal constraints when the required grasp points are inaccessible initially.

Previous approaches have often addressed the reorientation problem in a single-arm robotic setup. These methods typically rely on external fixtures ~\cite{luo2025fmb, kiyokawa2025soft} or require finding a stable intermediate pose on a planar surface ~\cite{wada2022reorientbot, mishra2024reorientdiff}, which inherently limits the flexibility of manipulation. In contrast, we tackle the reorientation problem within a collaborative dual-arm framework, mirroring the way humans naturally use both hands to perform such tasks (as shown in Figure \ref{fig:teaser}): one arm grasps the object, rotates it to an intermediate pose, and then hands it off to the other arm, which proceeds to place the object into the desired target pose. 

While a dual-arm configuration offers greater flexibility in selecting intermediate poses by mitigating gravitational constraints, it simultaneously introduces challenges in balancing trade-offs among grasp selection, trajectory length, and intermediate pose choice. Prior reorientation approaches typically decompose object rearrangement tasks into sequential stages, relying on rejection sampling ~\cite{wada2022reorientbot} or heuristics to generate solutions. However, such methods require a large number of candidate random samples and ultimately limit efficiency and overall performance.

To address these challenges, we propose a unified optimization framework for bimanual object reorientation that jointly identifies feasible manipulation trajectories and enables efficient execution. Specifically, our method integrates grasp quality, motion planning, collision avoidance, and handover feasibility. These constraints are encoded as differentiable energy-based models (EBMs)~\cite{Tomczak2024} and combined with diffusion scores from a pretrained grasp model. The framework then employs gradient-guided MCMC sampling \cite{du2023reduce} to generate bimanual grasp poses from the constrained posterior, effectively balancing the competing objectives. We evaluate our method across various simulated scenarios, and the experimental results show that our approach outperforms baseline methods by a substantial margin. Our contributions can be summarized as follows:
\begin{itemize}
\item We propose a \textbf{bimanual object reorientation framework} that \textbf{compositionally optimizes} grasp quality alongside motion plans during the reverse diffusion process, while proactively avoiding inter-arm and scene collisions.
\item This capability is powered by a \textbf{learning-based inverse kinematics model} we developed, which supports swift evaluation of motion costs and feasibility.
\item The efficacy of our approach is rigorously validated through extensive tests in both simulated and real-world domains, revealing its strong generalization across diverse scenes and its capacity to dramatically reduce joint movements.
\end{itemize}

\section{RELATED WORKS}

\paragraph{Object Reorientation for Rearrangement}
Conventional works on object rearrangement are typically studied within the Task and Motion Planning (TAMP) framework. Methods such as PDDLStream~\cite{garrett2020pddlstream}, LOFT~\cite{silver2021learning}, and Wells et al.~\cite{wells2019learning} adopt a modular architecture that combines symbolic planning with sampling-based motion planners (e.g., RRT) and heuristic grasp sampling. These approaches generally assume fully observable states and known object models, which limits their applicability in real-world open environments.

More recent methods leverage vision-language models to infer target object poses from language or image goals~\cite{kapelyukh2023dall, xu2024grasp, kapelyukh2024dream2real}, followed by pick-and-place execution. EBM-based approaches~\cite{gkanatsios2023energy, xu2024set} further model constraints to generate feasible goal poses. However, both traditional TAMP and recent learning-based pipelines often assume compatible grasps for pick-and-place stages and struggle with tight integration between grasp selection and downstream motion feasibility.

When direct placement is infeasible, robots must \emph{reorient} the object via intermediate poses and regrasps. Prior reorientation methods generate such poses through rejection sampling~\cite{wada2022reorientbot}, imitation learning~\cite{luo2025fmb}, or classifier-guided sampling from learned priors~\cite{mishra2024reorientdiff}. Nevertheless, these works are predominantly single-arm and do not jointly optimize reorientation poses with full motion trajectories and efficiency. 

\paragraph{Bimanual Manipulation}
Bimanual manipulation supports complex tasks such as fixture-based stabilization~\cite{wang2025learning}, cooperative carrying~\cite{zhang2025adaptive},
and object rearrangement~\cite{zhang2024learning, li2023efficient}.
Existing rearrangement approaches often decompose the problem into discrete stages (e.g., arm scheduling~\cite{zhang2024learning} or handover planning~\cite{li2023efficient}),
which can yield kinematically infeasible transitions or suboptimal reorientation due to heuristic grasp sampling that ignores full trajectory cost.

\paragraph{Energy-based Models and Compositional Optimization}
EBMs~\cite{lecun2006tutorial} provide a general energy-minimization view of control and planning.
Recent work connects diffusion models to EBMs~\cite{yang2023diffusion}, enabling composition of multiple diffusion models~\cite{du2019implicit, du2023reduce}
and additional constraints~\cite{huang2025hgdiffuser} for compositional generalization.
In particular, SE(3)-Diffusion~\cite{urain2023se} demonstrates the power of this paradigm by unifying grasp generation and trajectory planning within a single EBM framework. Building upon this line of work, we extend compositional EBMs to the more challenging setting of bimanual object reorientation. 

In contrast to prior modular or single-arm approaches, we treat bimanual reorientation as a \emph{unified} energy-based optimization problem. Using compositional EBMs and grasp diffusion models, we jointly optimize grasps, handover configurations, and smooth dual-arm motions, enabling tighter integration of constraints such as dual-arm coordination, collision avoidance, and grasp quality across the entire reorientation sequence.

\section{PRELIMINARIES}

In this work, we leverage diffusion models as the core generative component for grasp candidate initialization and subsequent joint refinement with motion planning. 
Diffusion models are particularly well-suited for this setup due to their close connection to energy-based models (EBMs) and their ability to support compositional optimization in high-dimensional, multi-modal spaces~\cite{du2023reduce,yang2023diffusion}.

Diffusion models \cite{ho2020denoising, song2020score} have emerged as powerful generative models capable of producing high-quality samples by learning to reverse a gradual noising process. 
In this work, we build upon the Denoising Diffusion Probabilistic Model (DDPM) framework \cite{ho2020denoising}, which has been successfully adapted to 6-DoF grasp generation tasks \cite{urain2023se, murali2025graspgen}.

Given a data distribution $q(\mathbf{x}_0)$, the forward diffusion process gradually adds Gaussian noise over $T$ timesteps according to a fixed variance schedule $\beta_1, \dots, \beta_T$:
\begin{equation}
q(\mathbf{x}_t \mid \mathbf{x}_{t-1}) = \mathcal{N}(\mathbf{x}_t; \sqrt{1-\beta_t} \mathbf{x}_{t-1}, \beta_t \mathbf{I}),
\label{eq:forward}
\end{equation}
where $\mathbf{x}_t \in \mathbb{R}^d$ is the noisy sample at timestep $t$. 
This process can be expressed in closed form as
\begin{equation}
q(\mathbf{x}_t \mid \mathbf{x}_0) = \mathcal{N}(\mathbf{x}_t; \sqrt{\bar{\alpha}_t} \mathbf{x}_0, (1 - \bar{\alpha}_t) \mathbf{I}),
\label{eq:closed_form}
\end{equation}
with $\alpha_t = 1 - \beta_t$ and $\bar{\alpha}_t = \prod_{s=1}^t \alpha_s$.

The reverse process is parameterized by a learned neural network $\epsilon_\theta(\mathbf{x}_t, t)$ that predicts the noise added at each step. 
The model is trained by minimizing the simplified variational lower bound, which reduces to the following denoising objective:
\begin{equation}
\mathcal{L}(\theta) = \mathbb{E}_{t,\mathbf{x}_0,\boldsymbol{\epsilon}} \Bigl[
\|\boldsymbol{\epsilon} - \epsilon_\theta(\sqrt{\bar{\alpha}_t} \mathbf{x}_0 + \sqrt{1-\bar{\alpha}_t} \boldsymbol{\epsilon}, t)\|^2
\Bigr],
\label{eq:loss_ddpm}
\end{equation}

During inference, sampling starts from pure noise $\mathbf{x}_T \sim \mathcal{N}(\mathbf{0},\mathbf{I})$ and iteratively denoises using
\begin{equation}
\mathbf{x}_{t-1} = \frac{1}{\sqrt{\alpha_t}} \Bigl( \mathbf{x}_t - \frac{1 - \alpha_t}{\sqrt{1 - \bar{\alpha}_t}} \epsilon_\theta(\mathbf{x}_t, t) \Bigr) + \sigma_t \mathbf{z}, \quad \mathbf{z} \sim \mathcal{N}(\mathbf{0},\mathbf{I}),
\label{eq:sampling}
\end{equation}
where $\sigma_t$ is typically set following the DDPM or DDIM \cite{song2020denoising} scheduler.

In grasp generation, the data $\mathbf{x}_0$ usually represents a grasp configuration (e.g., 6-DoF pose in SE(3)) conditioned on the object geometry (point cloud or other representations). 
GraspGen \cite{murali2025graspgen} adopts a DiffusionTransformer architecture to model this conditional distribution directly in grasp space, achieving state-of-the-art performance on diverse embodiments and clutter scenarios. 
Our framework initializes grasp candidates using a pretrained GraspGen model and subsequently refines them jointly with the motion plan through multi-objective optimization.

\section{METHODS}

\subsection{PROBLEM FORMULATION}

We study coordinated dual-arm pick-and-place in clutter using two UR12e 6-DoF robots. The robot must optimize a full reorientation sequence: pick with one arm, handover, regrasp with the other arm, and place at the target.

To explore the multi-modal grasp space and improve robustness, we optimize a batch of grasp tuples~$\mathcal{B} = \{\mathbf{g}^{(i)}\}_{i=1}^{N}$, 
where each~$\mathbf{g}^{(i)} = (g_{PK}^{(i)}, g_{HO}^{(i)}, g_{RG}^{(i)}, g_{PL}^{(i)}) \in SE(3)^4$ 
is an ordered quadruplet of 6-DoF world-frame poses: pick grasp ($g_{PK}$), handover pose ($g_{HO}$), regrasp pose ($g_{RG}$), and place grasp ($g_{PL}$). The ordered pairing ensures each candidate is evaluated as a consistent full manipulation sequence.

For each pair~$i$, we directly refine $g_{PK}^{(i)}$, $g_{HO}^{(i)}$, and $g_{PL}^{(i)}$, 
while $g_{RG}^{(i)} \in SE(3)$ is deterministically computed by a differentiable function~$h(\cdot)$ 
from~$g_{PK}^{(i)}$,~$g_{HO}^{(i)}$,~$g_{PL}^{(i)}$, the initial object pose, and the target object pose. 

Let~$\mathbf{P}_{\text{initial}}$ be the object point cloud at the initial pose, 
$\mathbf{P}_{\text{target}}$ the object point cloud at the target pose, 
and~$\mathbf{P}_{\text{scene}}$ the full scene point cloud. We solve for the best grasp tuple~$\mathbf{g}^* \in \mathcal{B}$ by minimizing a composite objective combining a pretrained diffusion prior and task constraints:
\begin{equation}
\resizebox{0.9\linewidth}{!}{$
\mathbf{g}^* = \arg\min_{\mathbf{g} \in \mathcal{B}} \Bigl[
\mathcal{L}_{\text{diff}}(g_{PK}, g_{PL} \mid \mathbf{P}_{\text{initial}}, \mathbf{P}_{\text{target}})
+ \sum_{c} \lambda_c \, \mathcal{L}_c(\mathbf{g}, \mathbf{P}_{\text{scene}})
\Bigr]
$},
\label{eq:composite_loss}
\end{equation}
where we optimize over the batch to retain diversity and escape local minima.

$\mathcal{L}_{\text{diff}}$ is the negative log-likelihood under the diffusion prior, conditioned on $g_{PK}$ and $g_{PL}$ (in the object frame) and the object point clouds $\mathbf{P}_{\text{initial}}$ and $\mathbf{P}_{\text{target}}$. The constraint terms $\mathcal{L}_c(\mathbf{g}, \mathbf{P}_{\text{scene}})$ depend on the full tuple and include joint-space smoothness $\mathcal{L}_{\text{smooth}}$ (via IK joint configurations $\mathbf{q}$), regrasp feasibility/safety $\mathcal{L}_{\text{regrasp}}$, handover feasibility $\mathcal{L}_{\text{handover}}$, and scene collision penalties $\mathcal{L}_{\text{collision}}$ w.r.t.\ $\mathbf{P}_{\text{scene}}$.

The selected tuple~$\mathbf{g}^*$ is then converted by inverse kinematics to waypoints $\mathbf{q}_{PK}$, $\mathbf{q}_{HO}$, $\mathbf{q}_{RG}$, $\mathbf{q}_{PL} \in \mathbb{R}^6$, which a motion planner uses to generate the dual-arm trajectory~$\tau$.

\subsection{OVERVIEW}
\begin{figure*}[!ht]
    \centering
    \includegraphics[trim=30 50 30 50, clip, width=0.7\textwidth]{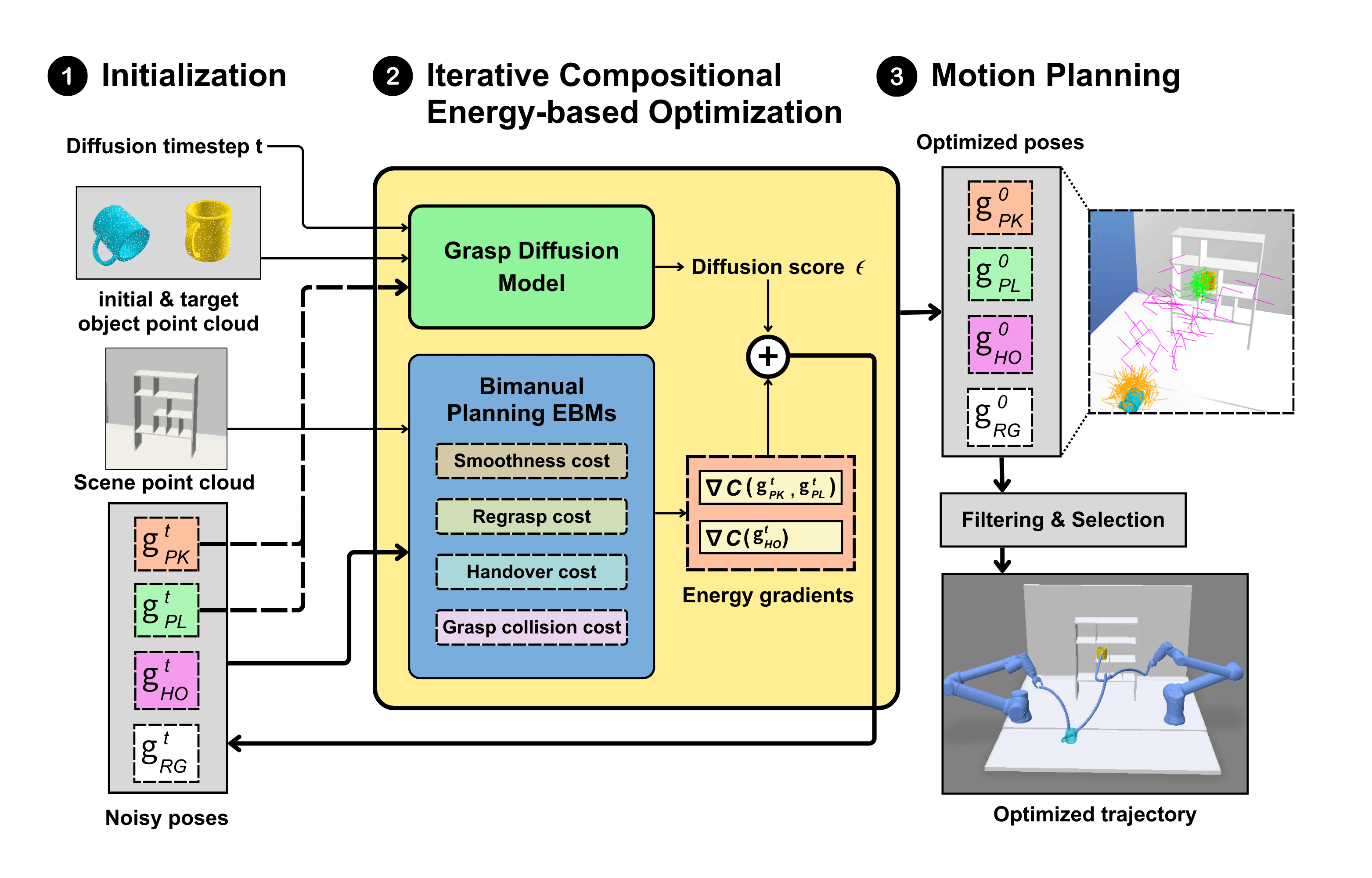}
    \caption{Overview of the proposed \textbf{BiCompoDiff} framework pipeline. 
    \textbf{Phase 1:} Noisy initial poses ($g_{PK}^{t}$, $g_{PL}^{t}$, $g_{HO}^{t}$ in world frame) are randomly sampled, and $g_{RG}^{t}$ is derived from the other three. 
    Point clouds (object + environment) are fed into the grasp diffusion model and the bimanual planning EBMs. 
    \textbf{Phase 2:} The diffusion model predicts $\epsilon$-scores with current timestep $t$ and noisy poses ($g_{PK}^{t}$, $g_{PL}^{t}$ converted to object frame), which are combined with energy gradients from the compositional bimanual planning EBMs (encoding smoothness, regrasp, handover, and grasp collision costs). 
    These guidance signals iteratively refine the noisy poses ($g_{PK}^{t}$, $g_{PL}^{t}$, $g_{HO}^{t}$) at each reverse diffusion timestep. 
    \textbf{Phase 3:} At timestep $t=0$, collision-free pose pairs are filtered and ranked by composite scores. 
    The highest-scoring pair is then passed to the sandbox module to generate the whole bimanual reorientation trajectory.
    }
    \label{fig:pipeline}
\end{figure*}

The overall pipeline is illustrated in Fig.~\ref{fig:pipeline}. 
Our framework, \textbf{Bimanual Compositional Diffusion and Energy-based Optimization, (BiCompoDiff)} jointly optimizes 6-DoF grasp poses for dual-arm pick-and-place tasks in cluttered environments by integrating a pretrained grasp diffusion model with compositional energy-based guidance during the reverse diffusion process. 
This enables the generation of constraint-aware, high-quality grasp sequences (pick, handover, regrasp, place) without retraining the generative model. 
The approach consists of three tightly coupled components:

\begin{enumerate}
    \item \textbf{Grasp Diffusion Model:} \quad 
    We utilize a pretrained grasp generation model (GraspGen~\cite{murali2025graspgen}) that has learned a rich multi-modal distribution over valid 6-DoF grasps conditioned on object geometry, to refine $\mathbf{g}_{Pk}^{t}$ and $\mathbf{g}_{PL}^{t}$ under reverse diffusion.
    
    \item \textbf{Bimanual Planning EBMs:} \quad
    To steer the denoising steps toward feasible bimanual manipulation, we define a collection of differentiable task-specific energy functions whose negative gradients serve as compositional guidance signals with the diffusion model's learned score function in each denoising step.

    \item \textbf{Compositional Annealed MCMC Sampling:} \quad 
    The reverse diffusion process is realized via annealed MCMC sampling, where the total energy landscape is the sum of the pretrained grasp diffusion energy and the planning energy terms. 
\end{enumerate}

\subsection{Grasp Diffusion Model}
We initialize grasps using the pretrained GraspGen model~\cite{murali2025graspgen}, a DiffusionTransformer that generates multi-modal 6-DoF grasp pose distributions conditioned on object point clouds. We use the \emph{graspgen\_robotiq\_2f\_140} checkpoint for inference. 

At inference, we run guided reverse diffusion for a fixed $k$ denoising steps. To support dual-arm pick-and-place, we condition sampling on both $\mathbf{P}_{\text{initial}}$ and $\mathbf{P}_{\text{target}}$, producing paired candidates in one pass: $\mathbf{g}_{Pk}^{t}$ (conditioned on $\mathbf{P}_{\text{initial}}$) and $\mathbf{g}_{PL}^{t}$ (conditioned on $\mathbf{P}_{\text{target}}$), keeping both within the learned grasp manifold. We also use GraspGen's discriminator to assign a scalar grasp confidence score to each generated (and optimized) candidate. This score measures consistency with high-probability regions of the pretrained distribution and serves as a distribution-aware term in our composite scoring function for selecting the optimal solution after optimization.

\subsection{Bimanual Planning EBMs}
These EBMs provide differentiable gradients that steer reverse diffusion toward grasp poses satisfying bimanual constraints. Gradients are taken w.r.t.\ the grasp pair $\mathbf{g}_{Pk}^{t}$, $\mathbf{g}_{HO}^{t}$, $\mathbf{g}_{RG}^{t}$, and $\mathbf{g}_{PL}^{t}$. We define task energies whose negative gradients act as corrective forces during denoising:

\textbf{Grasp collision avoidance} repels the gripper from obstacles. We approximate the gripper by collision spheres (transformed to world per pose) and query signed distances to the scene point cloud $\mathbf{P}_{\text{scene}}$ (KDTree). A soft-min aggregates the closest approach, and we apply a quadratic penalty when below margin $\delta$:
$c_{\text{coll}} = \frac{v^2}{\delta^2}, \quad v = \max(0, \delta - d_{\min})$,
yielding smooth repelling gradients near obstacles.

\textbf{Trajectory smoothness} minimizes joint motion between key configurations using a wrapped L1 proxy:
\begin{equation}
c_{\text{smooth}} = \|\Delta \mathbf{q}_{\text{PK}\to\text{HO}}\|_1 + \|\Delta \mathbf{q}_{\text{RG}\to\text{PL}}\|_1,
\end{equation}
This serves as a computationally efficient proxy for total joint travel under linear joint-space interpolation. We verified its strong correlation with the actual total joint angle change of cuRobo-generated trajectories across scenes of varying collision complexity. While analytical IK and other IK solvers \cite{sundaralingam2023curoboar} are accurate, it is discrete/non-differentiable and offers no guidance for unreachable targets. We therefore use a trained differentiable model, Joint-Decoupled SubnetMLP (SubnetIK) for the UR12e robot. For each key pose $\mathbf{g}_k$ ($k \in \{\text{PK}, \text{HO}, \text{RG}, \text{PL}\}$), SubnetIK predicts 8 joint candidates $\mathbf{q}_k^j$ ($j=1,\dots,8$); valid solutions are filtered by FK pose error threshold, and we select the four-pose joint candidate set minimizing $c_{\text{smooth}}$.

Analytical IK would require costly, unstable finite differences to approximate $\partial c_{\text{smooth}} / \partial \mathbf{g}_k$, especially near singularities or unreachable poses. In contrast, SubnetIK provides a smooth IK approximation $\mathbf{q}_k \approx f_{\text{SubnetIK}}(\mathbf{g}_k)$ with exact gradients via the chain rule: {\small $\dfrac{\partial c_{\text{smooth}}}{\partial \mathbf{g}_k} = \dfrac{\partial c_{\text{smooth}}}{\partial \mathbf{q}_k} \cdot J_f(\mathbf{g}_k)$}, where $J_f = \partial f_{\text{SubnetIK}} / \partial \mathbf{g}_k$ is the network Jacobian (autograd). SubnetIK is trained from analytical $\mathbf{q}$; for unreachable poses, the closest reachable pose provides supervision, yielding meaningful continuous gradients and stable optimization. A comparison to analytical IK is in Table \ref{tab:ik_comparison} (Appendix).

\textbf{Handover feasibility and safety} ensures that the handover pose $\mathbf{g}_{HO}^{t}$ is kinematically feasible and collision-free. 
Feasibility is enforced with a soft workspace bounding box penalty on position and the following orientation cost:
\begin{equation}
c_{\text{orient}} = \max(-\mathbf{d}^T \mathbf{R} \mathbf{e}_z, 0) + \max(c_{\min} - |(\mathbf{R} \mathbf{e}_z)_x|, 0),
\end{equation}
where $\mathbf{R}$ is the gripper rotation matrix at $\mathbf{g}_{HO}$, $\mathbf{d}=[-1,0,0]^\top$ is the desired outward direction, and $\mathbf{e}_z=[0,0,1]^\top$ is the gripper's local z-axis. The first term penalizes inward-pointing gripper orientations, while the second term penalizes near-planar alignments of the gripper z-axis with the robot's $xz$-plane ($c_{\min}$ controls the minimum allowed magnitude). 
Safety is enforced using signed-distance field penalties from collision spheres at the handover pose to scene obstacles, mirroring the grasp collision cost.

\textbf{Regrasp safety} prevents unsafe proximity between the pick and placement grippers during regrasp. Both $\mathbf{g}_{Pk}^{t}$ and $\mathbf{g}_{PL}^{t}$ are projected onto the initial object frame; collision spheres are placed accordingly, and the regrasp cost penalizes insufficient clearance:
\begin{equation}
c_{\text{regrasp}} = \left( \max(d_{\text{safe}} - d_{\min}, 0) / d_{\text{safe}} \right)^2
\end{equation}
where $d_{\min}$ is the minimum surface-to-surface distance between any pair of spheres from the two grippers, and $d_{\text{safe}}$ is the minimum required separation.

\subsection{Iterative Compositional Energy-Based Optimization with Annealed MCMC}
\begin{algorithm}[!ht]
\scriptsize
\caption{Iterative Compositional Energy-Based Optimization with Annealed MCMC}
\label{alg:iterative-mcmc-sampling}
\begin{algorithmic}[1]
\Require
    Initial noise $\mathbf{g}_{\text{PK}}^T, \mathbf{g}_{\text{PL}}^T \sim \mathcal{N}(0,I)$ \\
    Handover pose $\mathbf{g}_{\text{HO}}^T$ sampled from a predefined handover region \\
    Timesteps $T$, MCMC steps per timestep $M$ \\
    Denoising network $\varepsilon_\theta(\cdot, t; \mathbf{P}_{\text{initial}}, \mathbf{P}_{\text{target}})$ \\
    Scene point cloud $\mathbf{P}_{\text{scene}}$ \\
    Initial and target object point clouds $\mathbf{P}_{\text{initial}}$, $\mathbf{P}_{\text{target}}$ \\
    Planning energy functions $c_i(\cdot)$, weights $\lambda_i$ \\
    DDPM scheduling terms $\alpha_t$, $\bar{\alpha}_t$, $\sigma_t$

\State \quad   

\For{$t \gets T$ \textbf{down to} $1$} \Comment{Denoising loop over diffusion timesteps}
    \State \quad   

    \For{$m \gets 1$ \textbf{to} $M$} \Comment{MCMC refinement steps at timestep $t$}
    
        \State $\bm{\mu}_{\text{PK}}^t \gets \frac{1}{\sqrt{\alpha_t}} \Bigl( \mathbf{g}_{\text{PK}}^t - \frac{1-\alpha_t}{\sqrt{1-\bar{\alpha}_t}} \varepsilon_\theta(\mathbf{g}_{\text{PK}}^t, \mathbf{g}_{\text{PL}}^t, t; \mathbf{P}_{\text{initial}}, \mathbf{P}_{\text{target}}) \Bigr)$
        \State $\bm{\mu}_{\text{PL}}^t \gets \frac{1}{\sqrt{\alpha_t}} \Bigl( \mathbf{g}_{\text{PL}}^t - \frac{1-\alpha_t}{\sqrt{1-\bar{\alpha}_t}} \varepsilon_\theta(\mathbf{g}_{\text{PK}}^t, \mathbf{g}_{\text{PL}}^t, t; \mathbf{P}_{\text{initial}}, \mathbf{P}_{\text{target}}) \Bigr)$
        
        \State $\mathbf{g}_{\text{RG}}^t \gets h\bigl( \bm{\mu}_{\text{PK}}^t, \bm{\mu}_{\text{PL}}^t, \mathbf{g}_{\text{HO}}^t \bigr)$ \Comment{Regrasp pose (deterministic)}
        
        \State $\bm{\nabla g} \gets -\sum_i \lambda_i \nabla_{\mathbf{g}} \mathcal{C}_i\bigl( \bm{\mu}_{\text{PK}}^t, \bm{\mu}_{\text{PL}}^t, \mathbf{g}_{\text{HO}}^t, \mathbf{g}_{\text{RG}}^t; \mathbf{P}_{\text{scene}} \bigr)$ \Comment{task gradients}
        
        \If{$m = M$} \Comment{predict previous sample}
            \State $\mathbf{g}_{\text{PK}}^{t-1} \gets \bm{\mu}_{\text{PK}}^t + \bm{\nabla g}_{\text{PK}} + \sigma_t \mathbf{z}_{\text{PK}}$, \quad $\mathbf{z}_{\text{PK}} \sim \mathcal{N}(0,I)$
            \State $\mathbf{g}_{\text{PL}}^{t-1} \gets \bm{\mu}_{\text{PL}}^t + \bm{\nabla g}_{\text{PL}} + \sigma_t \mathbf{z}_{\text{PL}}$, \quad $\mathbf{z}_{\text{PL}} \sim \mathcal{N}(0,I)$
            \State $\mathbf{g}_{\text{HO}}^{t-1} \gets \mathbf{g}_{\text{HO}}^t + \bm{\nabla g}_{\text{HO}}$ 
        \Else \Comment{Refine current sample}
            \State $\mathbf{g}_{\text{PK}}^t \gets \bm{\mu}_{\text{PK}}^t + \bm{\nabla g}_{\text{PK}} + \sigma_t \mathbf{z}_{\text{PK}}$
            \State $\mathbf{g}_{\text{PL}}^t \gets \bm{\mu}_{\text{PL}}^t + \bm{\nabla g}_{\text{PL}} + \sigma_t \mathbf{z}_{\text{PL}}$
            \State $\mathbf{g}_{\text{HO}}^t \gets \mathbf{g}_{\text{HO}}^t + \bm{\nabla g}_{\text{HO}}$ 
        \EndIf
        
    \EndFor
\EndFor

\State \Return $\mathbf{g}_{\text{PK}}^0$, $\mathbf{g}_{\text{PL}}^0$, $\mathbf{g}_{\text{HO}}^0$ \Comment{Final optimized pick, place, handover poses}
\end{algorithmic}
\end{algorithm}


We optimize grasp poses via annealed MCMC-guided reverse diffusion (Alg.~\ref{alg:iterative-mcmc-sampling}) with default $M{=}5$.
We define \textbf{$k$} as the number of final denoising steps where bimanual planning EBM gradients are applied. 
For timesteps $t > k$, we perform standard reverse diffusion without task constraints and MCMC. 
For each timestep $t \leq k$, we run $M$ MCMC refinements: the diffusion model predicts the mean $\bm{\mu}^t$, we add weighted task constraint gradients, and inject noise. 
Only the final ($M$-th) refinement advances to timestep $t-1$. Inspired by inference-time policy steering~\cite{wang2025inference}, this loop better approximates sampling from the constrained posterior, yielding grasps that satisfy both diffusion likelihood and task constraints. This compositional guidance drives sampling toward feasible, smooth, and collision-free bimanual motions.

After optimization, the candidate pairs are first filtered by the cuRobo \cite{sundaralingam2023curoboar} collision checker . 
The remaining candidates are then ranked using a composite score that balances grasp quality, trajectory smoothness, and regrasp safety:
\begin{equation}
\text{composite} = w_{\text{conf}} \cdot \text{conf}_{\text{norm}} + w_{\text{plan}} \cdot (1 - \text{planning}_{\text{norm}})
\end{equation}
where $\text{conf}_{\text{norm}}$ is the normalized average confidence score of $g_{\text{PK}}, g_{\text{PL}}$ from GraspGen, and $\text{planning}_{\text{norm}}$ is the normalized planning cost combining trajectory smoothness and regrasp cost ($w_{\text{conf}}=0.4$,  $w_{\text{plan}}=0.6$).
The pair with the highest composite score is selected as the final solution and passed to the motion planner.

\section{EXPERIMENTS}

\begin{figure}[t]
    \includegraphics[trim=0 0 0 0, clip,width=250px]{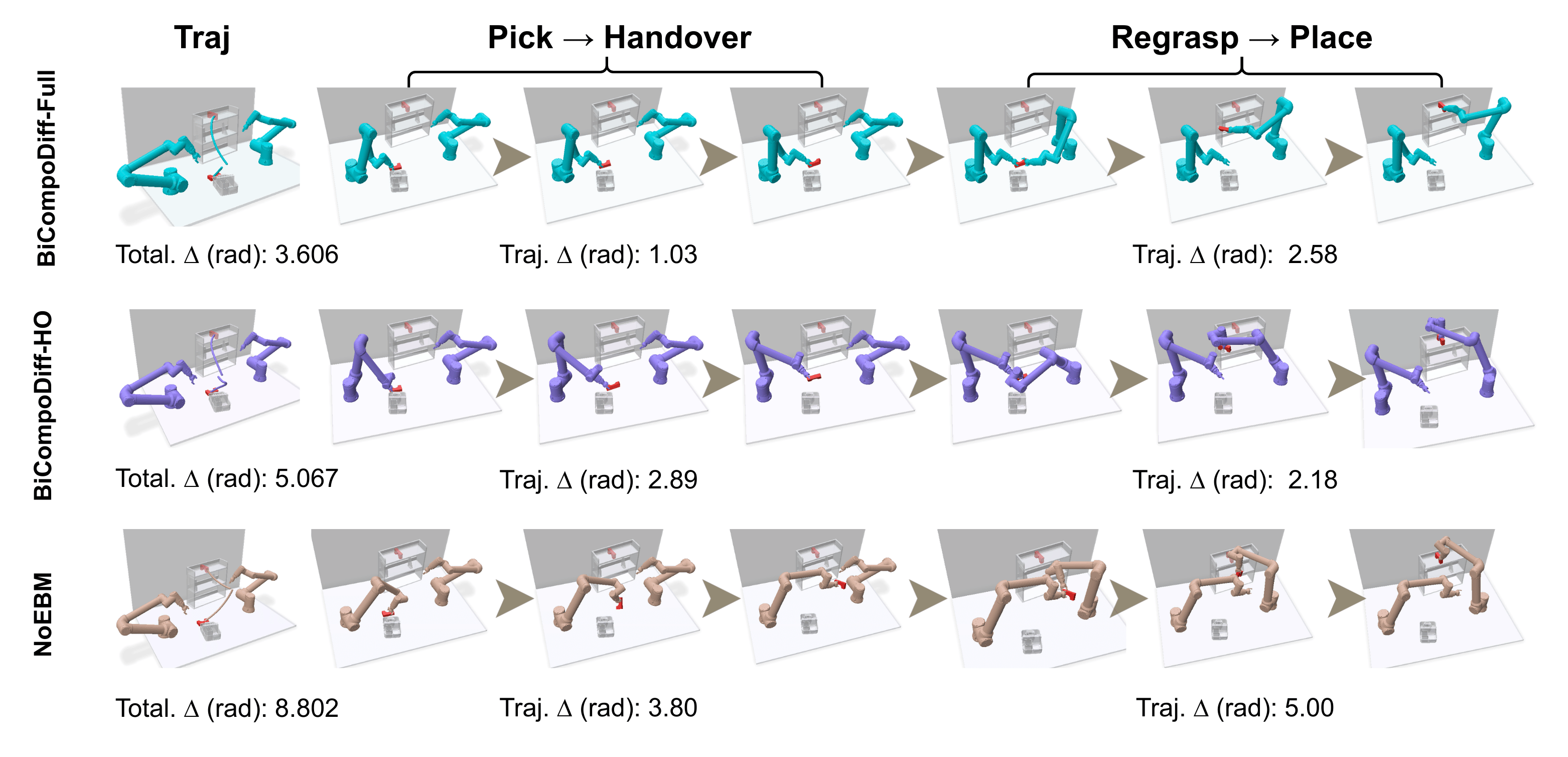}
    \caption{Visualization of bimanual reorientation trajectories for placing a bridge-shape brick from lying flat on the ground to standing upright on a shelf. \textbf{BiCompoDiff-Full} minimizes joint displacement from PK to HO while enabling smooth side-regrasping for final placement. \textbf{BiCompoDiff-HO} optimizes handover to a near-standing pose, reducing reorientation effort during the final RG to PL phase and yielding the most efficient RG to PL trajectory, but at the cost of higher joint delta earlier. \textbf{NoEBM} produces inefficient trajectories, requiring excessive object flipping and rear regrasp. Overall, \textbf{Full mode} achieves the best balance between PK to HO and RG to PL trajectory efficiency.}
    \label{fig:visualization}
\end{figure}
We evaluate BiCompoDiff on collaborative bimanual manipulation tasks involving pick, place, handover, and regrasp actions for object reorientation. All experiments are conducted in simulation unless otherwise stated, with real-world validation. 

\subsection{Implementation Details}
\label{subsec:impl_details}

\subsubsection{Simulation Pipeline}

Experiments are conducted in a modular simulation pipeline that couples a high-level task specification front-end with a motion planning back-end. The environment includes:
\begin{itemize}
    \item \textbf{Task Sandbox:} Defines scene geometry, initial/target object poses, and planner interface. Scenes are designed in Fusion 360 and imported as triangle meshes and point clouds ($\mathbf{P}_{\text{scene}}$, $\mathbf{P}_{\text{initial}}$, $\mathbf{P}_{\text{target}}$).
    \item \textbf{Simplified Physics \& Rendering:} Polyscope handles visualization. Grasping and placement are simulated by rigidly attaching/detaching the object mesh to the gripper at optimized key poses ($\mathbf{g}_{\text{PK}}$, $\mathbf{g}_{\text{HO}}$, $\mathbf{g}_{\text{RG}}$, $\mathbf{g}_{\text{PL}}$), abstracting low-level contact dynamics to focus on high-level planning.
    \item \textbf{Motion Planning:} cuRobo generates minimum-jerk, collision-free joint trajectories between key poses using the robot URDF, full scene geometry, and joint limits.
\end{itemize}

\subsubsection{Task Benchmark}

We introduce a benchmark of 60 pick-and-place reorientation tasks of varying difficulty (20 easy, 20 medium, 20 hard), all designed in Fusion 360. Quantitative results are averaged across the full set.
\begin{itemize}
    \item \textbf{Easy:} Minimal scene (ground plane + object in the air); 
    \item \textbf{Medium:} Static obstacles (e.g., shelf) added, requiring active collision avoidance during coordination.
    \item \textbf{Hard:} Dense, realistic clutter (e.g., bin picking + cluttered shelf placement); tests tight-space navigation and multi-constraint satisfaction.
\end{itemize}

\subsubsection{Evaluation Metrics}

Motion efficiency (excluding home-to-home segments) is measured by:
\begin{itemize}
    \item \textbf{Total joint-space displacement} (\textbf{Traj.~$\Delta$} in rad) -- the \textbf{primary metric} and focused optimization aspect,
    \item End-effector trajectory length (\textbf{Traj.~Len.} in m),
    \item Execution duration (\textbf{Dur} in s).
\end{itemize}
The latter two metrics are treated as secondary.
Grasp quality is quantified by pick and place confidence scores (\textbf{PK Conf.}, \textbf{PL Conf.}) and percentage of gripper finger pad contact area on the object (\textbf{PK CA \%}, \textbf{PL CA \%})~\cite{leddy2020frictional}. Regrasp cost (\textbf{RG Cost}) penalizes gripper-to-gripper collision during regrasp. Path success rate (\textbf{Succ.\%}) reports the percentage of fully collision-free trajectories (no robot-robot or robot-scene collisions), while the average number of collision-free grasp pairs (\textbf{CF}) assesses grasp collision avoidance. Computational cost is the average inference time per plan (see Appendix).

\subsubsection{Baselines}

We implement two strong baselines to isolate our contributions.

\textbf{NoEBM} follows the same pipeline as BiCompoDiff but ablates bimanual planning EBM gradients: pick ($\mathbf{g}_{\text{PK}}$) and place ($\mathbf{g}_{\text{PL}}$) poses are denoised solely using the GraspGen diffusion score function, without any bimanual planning EBM gradients.

\textbf{ReorientBot} adapts ReorientBot~\cite{wada2022reorientbot} to our bimanual setting. For comparison, both ReorientBot and BiCompoDiff use GraspGen to sample 100 pick grasps ($g_{PK}$), 100 place grasps ($g_{PL}$), and random handover grasps ($g_{HO}$). We replace learned filtering with transparent heuristic filtering: IK solutions $\mathbf{q}$ are computed for each pose, followed by (i) scene collision checking with cuRobo, (ii) joint-space trajectory validity scoring, and (iii) smoothness scoring via total L1 joint displacement. Success is evaluated on scene and self-collisions only (dual-arm inter-collisions ignored for fairness). The key difference is grasp pairing: BiCompoDiff uses ordered sampling (100 ordered pairs), while ReorientBot exhaustively combines all grasps ($100^3$ candidates) before filtering.

\subsubsection{Parameters}
We tuned several cost weights through extensive experimentation. The final values used in all reported experiments are listed in Table~\ref{tab:cost_weights}, which yielded the best overall performance.

\subsection{Ablation Studies}

We ablate three key factors of BiCompoDiff across all tasks: (i) optimization modes, (ii) MCMC-based guidance, and (iii) the number of gradient-application step $k$.

\subsubsection{Optimization modes}

We compare two configurations (averaged over $k=3$, with MCMC enabled):
\begin{itemize}
    \item \textbf{BiCompoDiff-Full}: Full joint optimization. All poses ($\mathbf{g}_{\text{PK}}$, $\mathbf{g}_{\text{PL}}$, $\mathbf{g}_{\text{HO}}$) receive smoothness gradients plus task-specific gradients.
    \item \textbf{BiCompoDiff-HO}: Handover-only mode. Smoothness gradients are applied \emph{only} to $\mathbf{g}_{\text{HO}}$; $\mathbf{g}_{\text{PK}}$ and $\mathbf{g}_{\text{PL}}$  receive other task gradients only.
\end{itemize}
\begin{table*}[t]
\vspace{4pt} 
\footnotesize
\centering
\setlength{\tabcolsep}{3.5pt}
\caption{Performance comparison across configurations (k=3, 200 grasp pairs).}
\label{tab:opt_modes_and_mcmc}
\begin{tabular}{@{} l *{10}{c} @{}}
\toprule
Config & Traj. $\Delta$ (rad)$\downarrow$ & Traj. Len. (m)$\downarrow$ & Dur (s)$\downarrow$ & PK Conf.$\uparrow$ & PL Conf.$\uparrow$ & RG Cost$\downarrow$ & PK CA (\%) $\uparrow$ & PL CA (\%) $\uparrow$ & Succ (\%) $\uparrow$ & CF $\uparrow$ \\
\midrule
NoEBM              & 7.293 & 1.476 & 1.621 & 0.891 & 0.875 & 1.625 & \textbf{79.7} & 70.8  & 41.7  & 59.8  \\
BiCompoDiff-Full   & \textbf{4.569} & 1.184 & \textbf{1.243} & 0.808 & 0.863 & 0.259 & 77.7  & \textbf{72.7} & 63.3  & 106   \\
BiCompoDiff-HO     & 4.692 & \textbf{1.183} & 1.286 & \textbf{0.863} & \textbf{0.878} & \textbf{0.208} & 76.0  & 67.6  & \textbf{70.0} & \textbf{107} \\
Full-NoMCMC         & 6.150 & 1.380 & 1.522 & 0.760 & 0.615 & 0.340 & 60.4  & 48.3  & 57.8  & 95.3  \\
\bottomrule
\end{tabular}
\end{table*}

Table~\ref{tab:opt_modes_and_mcmc} summarizes results. Both variants improve motion efficiency and safety while keeping grasp quality competitive. BiCompoDiff-Full yields the smoothest and shortest trajectories overall (lowest joint delta, length, and duration): joint delta $-37\%$, length $-20\%$, duration $-23\%$, and regrasp cost $-84\%$, while increasing success $+22\%$ (abs.) and collision-free pairs $+77\%$. This highlights the benefit of injecting bimanual smoothness gradients across all poses during denoising.

BiCompoDiff-HO attains the highest success (70.0\%) and lowest regrasp cost. This shows that optimizing pick/place in Full further improves efficiency, at the modest cost of slightly reduced grasp and regrasp quality (and thus success). Figure~\ref{fig:visualization} provides a qualitative comparison.

\subsubsection{Effect of enabling MCMC guidance (Full mode, fixed at $k=3$)}

Table~\ref{tab:opt_modes_and_mcmc} shows that enabling MCMC improves all metrics, with a moderate increase in inference time. We therefore use MCMC by default.

\subsubsection{Effect of the number of $k$ (Full mode + MCMC)}

Figure~\ref{fig:k-ablation}  (Appendix) shows that increasing $k$ from 1 to 10 improves trajectory quality, grasp confidence/contact area, and collision-free pairs, with the main gains up to $k\approx5$--6 and saturation thereafter. Success mildly peaks at low-to-mid $k$ before slightly dropping at very high $k$; regrasp cost has no consistent trend; inference time increases with $k$. We set $k=3$ by default for the best quality--efficiency trade-off.

\subsection{Comparison to ReorientBot}
\begin{table*}[t]
\small
\centering
\setlength{\tabcolsep}{3.5pt}
\caption{Comparison with adapted ReorientBot baseline (k=3, 100 grasp pairs)}
\label{tab:reorientbot}
\begin{tabular}{@{} l *{9}{c} @{}}
\toprule
Config & Traj. $\Delta$ (rad)$\downarrow$ & Traj. Len. (m)$\downarrow$ & Dur (s)$\downarrow$ & PK Conf.$\uparrow$ & PL Conf.$\uparrow$ & PK CA (\%)$\uparrow$ & PL CA (\%)$\uparrow$ & Succ (\%)$\uparrow$ & CF$\uparrow$ \\
\midrule
NoEBM & 7.372 & 1.399 & 1.637 & 0.804 & 0.847 & 78.8 & 79.4 & 71.7 & 29.8 \\
ReorientBot*\cite{wada2022reorientbot} & 6.434 & 1.302 & 1.530 & 0.655 & 0.719 & 72.6 & 71.1 & 58.3 & \textbf{479.3} \\
BiCompoDiff-Full & \textbf{5.628} & \textbf{1.275} & \textbf{1.423} & \textbf{0.846} & \textbf{0.869} & 77.0 & 71.4 & \textbf{81.7} & 52.7 \\
Full (\textbf{w/o $\nabla g_{\rm coll}$}) & 5.717 & 1.295 & 1.477 & 0.801 & 0.847 & \textbf{80.7} & \textbf{81.2} & 65.0 & 29.2 \\
\bottomrule
\end{tabular}
\end{table*}

Table~\ref{tab:reorientbot} compares our method (\textbf{BiCompoDiff-Full}) with an adapted \textbf{ReorientBot-style} baseline.
\textbf{BiCompoDiff-Full} achieves markedly better overall performance: higher success (\textbf{81.7\%} vs.\ 58.3\%, \textbf{+23.4} abs.),
smoother motions with smaller joint delta (\textbf{-12.5\%}), shorter trajectories (\textbf{-2.1\%}), lower duration (\textbf{-7.0\%}).

ReorientBot benefits from \emph{dynamic grasp pairing} and can improve collision-free planning over our ordered-pair baseline (NoEBM),
but it degrades grasp quality and incurs high computational cost due to evaluating many grasp candidates.

In contrast, our \emph{paired} formulation with \textbf{compositional energy-based optimization} avoids exhaustive enumeration while simultaneously improving success rate and efficiency. Importantly, this gain is not merely due to more collision-free grasp pairs. Comparing BiCompoDiff-Full with its variant without grasp collision gradients (\textbf{w/o $\nabla g_{\rm coll}$}) shows that the efficiency boost primarily stems from injecting differentiable bimanual smoothness gradients during denoising. Overall, end-to-end gradient guidance enables more reliable and efficient execution than traditional sample-and-filter pipelines, at substantially lower computational cost.

\subsection{Real-World Experiments}
\label{subsec:real_world}

We validate BiCompoDiff on a real-world bimanual platform mirroring our simulation setup: two UR12e 6-DoF arms equipped with Robotiq Hand-E parallel-jaw grippers. Trajectories are executed from pre-computed joint sequences.

Qualitative results (supplementary video) demonstrate smooth and collision-free bimanual coordination with BiCompoDiff-Full in challenging scenes. 
Quantitative evaluation on two representative scenes shows significant trajectory efficiency gains: the T-shape brick improves from 10.850\,rad (NoEBM) to 4.371\,rad (59.7\%$\uparrow$), and the cup improves from 10.126\,rad to 5.427\,rad (46.4\%$\uparrow$). 
These results indicate successful sim-to-real transfer under the assumption of perfect perception, highlighting the practical benefit of diffusion-guided compositional energy optimization for smoother and more reliable bimanual reorientation.

\section{CONCLUSIONS}

In this work, we introduced \textbf{BiCompoDiff}, a compositional diffusion and energy-based optimization framework for end-to-end bimanual object reorientation. By integrating GraspGen with an energy-guided joint denoising process and annealed MCMC sampling, BiCompoDiff jointly optimizes pick, handover, regrasp, and place under trajectory smoothness, collision avoidance, handover, and regrasp feasibility constraints. SubnetIK further enables differentiable inverse kinematics for effective gradient-based refinement. Experiments and real-robot results show higher success rates ($\mathbf{>}$20\%$\uparrow$) and more efficient trajectories (37\%$\uparrow$) than strong baselines, including an adapted ReorientBot pipeline. Overall, BiCompoDiff demonstrates that compositional diffusion with task energies enables scalable co-optimization of high-dimensional grasp and key poses that account for motion-level objectives beyond sampling-and-filtering approaches. 

Current limitations of BiCompoDiff include the assumption of perfect perception (known object poses and geometries) and indirect trajectory optimization via optimized grasps and key poses that incorporate motion planning objectives. Future work could explore direct trajectory optimization, robustness to noisy or partial observations, and more flexible task-level reasoning.



\section*{ACKNOWLEDGMENTS}
This research is supported by the Hong Kong ITF grant GHP/215/23SZ, Shenzhen Loop Area Institute under grant FPF10120250006, and BYD-HKUST Joint Lab Research Project under grant BYD26IS02.

\bibliographystyle{IEEEtran}
\bibliography{IEEEabrv,related}

@STRING{tog       = "ACM Transactions on Graphics"}

@article{luo2025fmb,
  title={Fmb: a functional manipulation benchmark for generalizable robotic learning},
  author={Luo, Jianlan and Xu, Charles and Liu, Fangchen and Tan, Liam and Lin, Zipeng and Wu, Jeffrey and Abbeel, Pieter and Levine, Sergey},
  journal={The International Journal of Robotics Research},
  volume={44},
  number={4},
  pages={592--606},
  year={2025},
  publisher={Sage Publications Sage UK: London, England}
}

@inproceedings{wada2022reorientbot,
  title={ReorientBot: Learning object reorientation for specific-posed placement},
  author={Wada, Kentaro and James, Stephen and Davison, Andrew J},
  booktitle={2022 International Conference on Robotics and Automation (ICRA)},
  pages={8252--8258},
  year={2022},
  organization={IEEE}
}

@article{kiyokawa2025soft,
  title={Soft Regrasping Tool Inspired by Jamming Gripper},
  author={Kiyokawa, Takuya and Hu, Zhengtao and Wan, Weiwei and Harada, Kensuke},
  journal={arXiv preprint arXiv:2509.13815},
  year={2025}
}

@inproceedings{mishra2024reorientdiff,
  title={Reorientdiff: Diffusion model based reorientation for object manipulation},
  author={Mishra, Utkarsh A and Chen, Yongxin},
  booktitle={2024 IEEE International Conference on Robotics and Automation (ICRA)},
  pages={10867--10873},
  year={2024},
  organization={IEEE}
}

@incollection{Tomczak2024,
title={Energy-based models},
  author={Tomczak, Jakub M},
  booktitle={Deep Generative Modeling},
  pages={143--158},
  year={2021},
  publisher={Springer}
}

@article{kapelyukh2023dall,
  title={Dall-e-bot: Introducing web-scale diffusion models to robotics},
  author={Kapelyukh, Ivan and Vosylius, Vitalis and Johns, Edward},
  journal={IEEE Robotics and Automation Letters},
  volume={8},
  number={7},
  pages={3956--3963},
  year={2023},
  publisher={IEEE}
}

@article{xu2024grasp,
  title={Grasp, see, and place: Efficient unknown object rearrangement with policy structure prior},
  author={Xu, Kechun and Zhou, Zhongxiang and Wu, Jun and Lu, Haojian and Xiong, Rong and Wang, Yue},
  journal={IEEE Transactions on Robotics},
  volume={41},
  pages={464--483},
  year={2024},
  publisher={IEEE}
}

@inproceedings{kapelyukh2024dream2real,
  title={Dream2real: Zero-shot 3d object rearrangement with vision-language models},
  author={Kapelyukh, Ivan and Ren, Yifei and Alzugaray, Ignacio and Johns, Edward},
  booktitle={2024 IEEE International Conference on Robotics and Automation (ICRA)},
  pages={4796--4803},
  year={2024},
  organization={IEEE}
}

@article{gkanatsios2023energy,
  title={Energy-based models are zero-shot planners for compositional scene rearrangement},
  author={Gkanatsios, Nikolaos and Jain, Ayush and Xian, Zhou and Zhang, Yunchu and Atkeson, Christopher and Fragkiadaki, Katerina},
  journal={arXiv preprint arXiv:2304.14391},
  year={2023}
}

@article{xu2024set,
  title={“Set it up”: Functional object arrangement with compositional generative models},
  author={Xu, Yiqing and Mao, Jiayuan and Li, Linfeng and Du, Yilun and Loz{\'a}no-P{\'e}rez, Tomas and Kaelbling, Leslie Pack and Hsu, David},
  journal={The International Journal of Robotics Research},
  pages={02783649251378198},
  year={2024},
  publisher={SAGE Publications Sage UK: London, England}
}

@inproceedings{zhang2024learning,
  title={Learning Dual-arm Object Rearrangement for Cartesian Robots},
  author={Zhang, Shishun and She, Qijin and Li, Wenhao and Zhu, Chenyang and Wang, Yongjun and Hu, Ruizhen and Xu, Kai},
  booktitle={2024 IEEE International Conference on Robotics and Automation (ICRA)},
  pages={7440--7446},
  year={2024},
  organization={IEEE}
}

@inproceedings{li2023efficient,
  title={Efficient bimanual handover and rearrangement via symmetry-aware actor-critic learning},
  author={Li, Yunfei and Pan, Chaoyi and Xu, Huazhe and Wang, Xiaolong and Wu, Yi},
  booktitle={2023 IEEE International Conference on Robotics and Automation (ICRA)},
  pages={3867--3874},
  year={2023},
  organization={IEEE}
}

@inproceedings{urain2023se,
  title={Se (3)-diffusionfields: Learning smooth cost functions for joint grasp and motion optimization through diffusion},
  author={Urain, Julen and Funk, Niklas and Peters, Jan and Chalvatzaki, Georgia},
  booktitle={2023 IEEE international conference on robotics and automation (ICRA)},
  pages={5923--5930},
  year={2023},
  organization={IEEE}
}

@inproceedings{huang2025hgdiffuser,
  title={HGDiffuser: efficient task-oriented grasp generation via human-guided grasp diffusion models},
  author={Huang, Dehao and Dong, Wenlong and Tang, Chao and Zhang, Hong},
  booktitle={2025 IEEE/RSJ International Conference on Intelligent Robots and Systems (IROS)},
  pages={19538--19545},
  year={2025},
  organization={IEEE}
}

@article{lecun2006tutorial,
  title={A tutorial on energy-based learning},
  author={LeCun, Yann and Chopra, Sumit and Hadsell, Raia and Ranzato, M and Huang, Fujie and others},
  journal={Predicting structured data},
  volume={1},
  number={0},
  year={2006}
}

@article{yang2023diffusion,
  title={Diffusion models: A comprehensive survey of methods and applications},
  author={Yang, Ling and Zhang, Zhilong and Song, Yang and Hong, Shenda and Xu, Runsheng and Zhao, Yue and Zhang, Wentao and Cui, Bin and Yang, Ming-Hsuan},
  journal={ACM computing surveys},
  volume={56},
  number={4},
  pages={1--39},
  year={2023},
  publisher={ACM New York, NY, USA}
}

@article{du2019implicit,
  title={Implicit generation and modeling with energy based models},
  author={Du, Yilun and Mordatch, Igor},
  journal={Advances in neural information processing systems},
  volume={32},
  year={2019}
}

@inproceedings{du2023reduce,
  title={Reduce, reuse, recycle: Compositional generation with energy-based diffusion models and mcmc},
  author={Du, Yilun and Durkan, Conor and Strudel, Robin and Tenenbaum, Joshua B and Dieleman, Sander and Fergus, Rob and Sohl-Dickstein, Jascha and Doucet, Arnaud and Grathwohl, Will Sussman},
  booktitle={International conference on machine learning},
  pages={8489--8510},
  year={2023},
  organization={PMLR}
}

@article{zhang2025adaptive,
  title={Adaptive coordinated impedance control for dual-arm robot symmetric bimanual tasks},
  author={Zhang, Yang},
  journal={Robotics and Autonomous Systems},
  volume={193},
  pages={105110},
  year={2025},
  publisher={Elsevier}
}

@article{wang2025learning,
  title={Learning to assemble with alternative plans},
  author={Wang, Ziqi and Liu, Wenjun and Wang, Jingwen and Vallat, Gabriel and Shi, Fan and Parascho, Stefana and Kamgarpour, Maryam},
  journal={ACM Transactions on Graphics (TOG)},
  volume={44},
  number={4},
  pages={1--16},
  year={2025},
  publisher={ACM New York, NY, USA}
}

@article{ho2020denoising,
  title={Denoising diffusion probabilistic models},
  author={Ho, Jonathan and Jain, Ajay and Abbeel, Pieter},
  journal={Advances in neural information processing systems},
  volume={33},
  pages={6840--6851},
  year={2020}
}

@article{song2020score,
  title={Score-based generative modeling through stochastic differential equations},
  author={Song, Yang and Sohl-Dickstein, Jascha and Kingma, Diederik P and Kumar, Abhishek and Ermon, Stefano and Poole, Ben},
  journal={arXiv preprint arXiv:2011.13456},
  year={2020}
}

@article{murali2025graspgen,
  title={Graspgen: A diffusion-based framework for 6-dof grasping with on-generator training},
  author={Murali, Adithyavairavan and Sundaralingam, Balakumar and Chao, Yu-Wei and Yuan, Wentao and Yamada, Jun and Carlson, Mark and Ramos, Fabio and Birchfield, Stan and Fox, Dieter and Eppner, Clemens},
  journal={arXiv preprint arXiv:2507.13097},
  year={2025}
}

@article{song2020denoising,
  title={Denoising diffusion implicit models},
  author={Song, Jiaming and Meng, Chenlin and Ermon, Stefano},
  journal={arXiv preprint arXiv:2010.02502},
  year={2020}
}

@inproceedings{wang2025inference,
  title={Inference-time policy steering through human interactions},
  author={Wang, Yanwei and Wang, Lirui and Du, Yilun and Sundaralingam, Balakumar and Yang, Xuning and Chao, Yu-Wei and P{\'e}rez-D’Arpino, Claudia and Fox, Dieter and Shah, Julie},
  booktitle={2025 IEEE International Conference on Robotics and Automation (ICRA)},
  pages={15626--15633},
  year={2025},
  organization={IEEE}
}

@article{leddy2020frictional,
  title={Examining the frictional behavior of primitive contact geometries for use as robotic finger pads},
  author={Leddy, Michael T and Dollar, Aaron M},
  journal={IEEE Robotics and Automation Letters},
  volume={5},
  number={2},
  pages={3137--3144},
  year={2020},
  publisher={IEEE}
}

@article{sundaralingam2023curoboar,
  author    = {Balakumar Sundaralingam and Siva Kumar Sastry Hari and Adam Fishman and Caelan Garrett and Karl Van Wyk and Valts Blukis and Alexander Millane and Helen Oleynikova and Ankur Handa and Fabio Ramos and Nathan Ratliff and Dieter Fox},
  title     = {cuRobo: Parallelized Collision-Free Minimum-Jerk Robot Motion Generation},
  journal   = {arXiv preprint arXiv:2310.17274},
  year      = {2023},
  url       = {https://arxiv.org/abs/2310.17274}
}

@inproceedings{garrett2020pddlstream,
  author    = {Caelan R. Garrett and Tom{\'a}s Lozano-P{\'e}rez and Leslie Pack Kaelbling},
  title     = {{PDDLStream}: Integrating Symbolic Planners and Blackbox Samplers via Optimistic Adaptive Planning},
  booktitle = {International Conference on Automated Planning and Scheduling (ICAPS)},
  year      = {2020}
}

@inproceedings{silver2021learning,
  author    = {Silver, Tom and Chitnis, Rohan and Tenenbaum, Joshua and Kaelbling, Leslie Pack and Lozano-P{\'e}rez, Tom{\'a}s},
  title     = {Learning Symbolic Operators for Task and Motion Planning},
  booktitle = {2021 IEEE/RSJ International Conference on Intelligent Robots and Systems (IROS)},
  year      = {2021},
  pages     = {3182--3189},
  doi       = {10.1109/IROS51168.2021.9635941}
}

@article{wells2019learning,
  author  = {Wells, Andrew M. and Dantam, Neil T. and Shrivastava, Anshumali and Kavraki, Lydia E.},
  title   = {Learning Feasibility for Task and Motion Planning in Tabletop Environments},
  journal = {IEEE Robotics and Automation Letters},
  volume  = {4},
  number  = {2},
  pages   = {1255--1262},
  year    = {2019},
  doi     = {10.1109/LRA.2019.2894861}
}
\appendix
\begingroup
\setlength{\abovedisplayskip}{3pt}
\setlength{\belowdisplayskip}{3pt}
\setlength{\abovedisplayshortskip}{2pt}
\setlength{\belowdisplayshortskip}{2pt}

\setlength{\parskip}{2pt}
\setlength{\floatsep}{4pt}
\setlength{\textfloatsep}{6pt}
\setlength{\intextsep}{4pt}
\setlength{\dbltextfloatsep}{6pt}
\setlength{\dblfloatsep}{4pt}

\subsection{Joint-Decoupled SubnetMLP Model (SubnetIK)}

\begin{figure}[!ht]
  \centering
 \includegraphics[width={0.8\linewidth}]{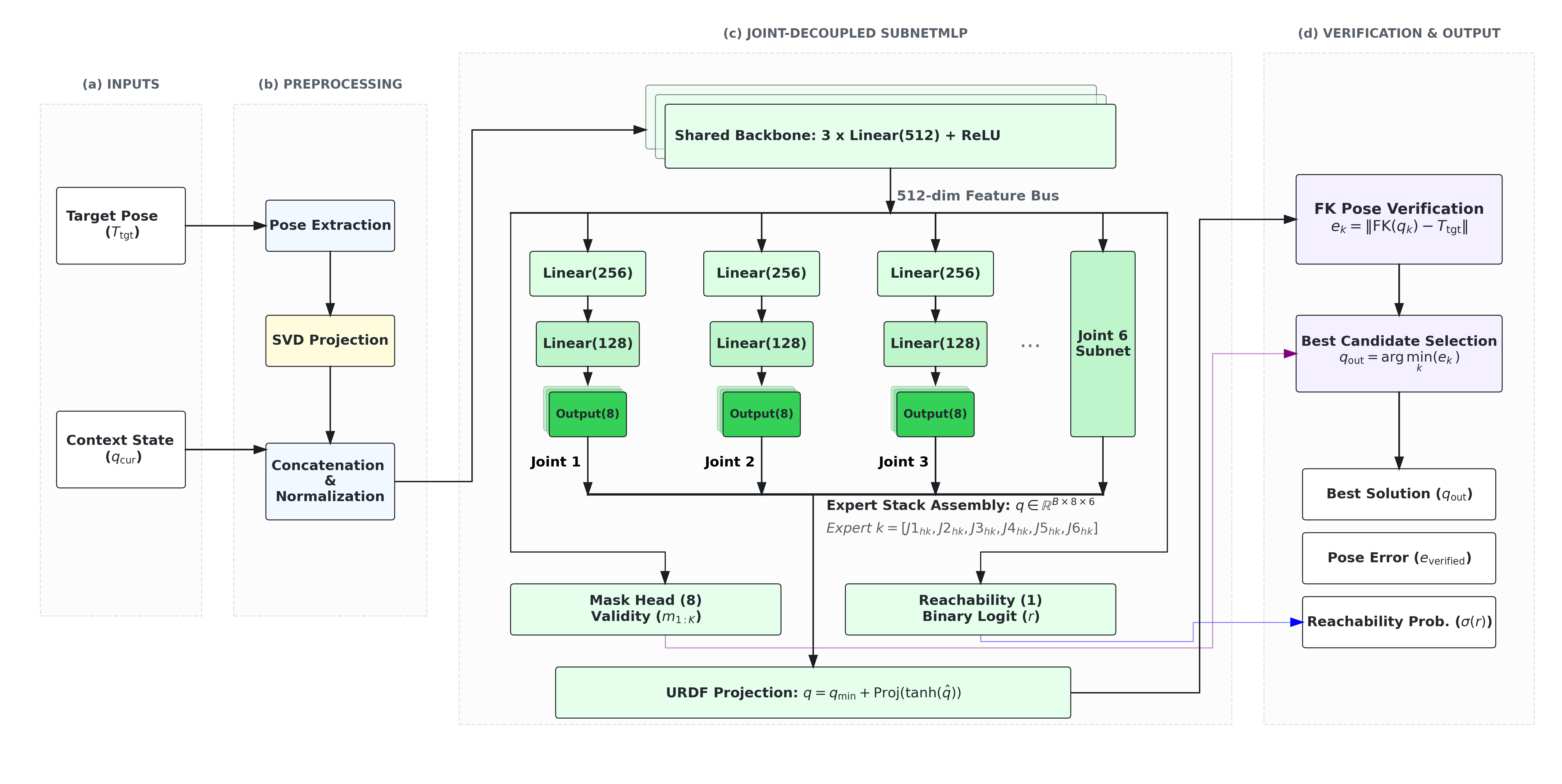}
  \caption{SubnetIK architecture}
  \label{fig:subnetik_workflow}
\end{figure}

SubnetIK (Figure~\ref{fig:subnetik_workflow}) takes as input a 9D pose representation (first two columns of the rotation matrix + translation vector) concatenated with the current joint state $\mathbf{q}_{\mathrm{cur}}$. 
The model consists of a shared backbone followed by six joint-specific sub-networks that predict per-joint hypotheses. 
The outputs are projected to satisfy URDF joint limits, and the model additionally predicts a reachability score and per-candidate validity logits.

\begin{table}[t]
\vspace{5pt}
\small
\caption{\textbf{IK comparison (mixed test set)} ($N{=}1000$; reachable + No-IK).
\textbf{Return\%}: queries returning a joint vector.
\textbf{Succ@10mm}: success rate where final pose error $\leq 10$\,mm. Non-returned counted as failures.
\textbf{MLP (Oracle)}: best of $K{=}8$ via FK; \textbf{MLP+FKRefine}: gradient refinement.}
\label{tab:appendix_comparison}
\centering
\footnotesize
\setlength{\tabcolsep}{4pt}
\renewcommand{\arraystretch}{1.08}

\begin{tabularx}{\columnwidth}{@{}lcccc@{}}
\toprule
Method & Return\%$\uparrow$ & Succ@10mm$\uparrow$ & Orient.\ (deg)$\downarrow$ & Time (ms)$\downarrow$ \\
\midrule
UR Analytic IK & 80.2 & 80.2 & 0.0001 & 0.031 \\
MLP (Oracle) & \textbf{100.0} & 85.1 & 3.35 & \textbf{0.067} \\
MLP+FKRefine & \textbf{100.0} & \textbf{90.1} & \textbf{2.11} & 0.341 \\
\bottomrule
\end{tabularx}
\end{table}

\begin{table}[!ht]
\centering
\small
\caption{\footnotesize Comparison of BiCompoDiff-Full ($k$=5) with Analytical IK, SubnetIK, and NoEBM baseline (average metrics)}
\label{tab:ik_comparison}
\setlength{\tabcolsep}{3.8pt}
\renewcommand{\arraystretch}{0.95}
\begin{tabular}{lccc}
\toprule
Metric & Analytical IK & SubnetIK & NoEBM \\
\midrule
Traj. $\Delta$ (rad) & 6.26 & \textbf{6.03} & 7.10 \\
Traj. Len. (m)       & 1.409 & 1.373 & \textbf{1.098} \\
Dur (s)              & 1.48  & \textbf{1.44}  & 1.54  \\
Infer. Time (s)      & 123.90 & \textbf{3.12}  & \textbf{0.11}  \\
\bottomrule
\end{tabular}
\end{table}

\subsection{Experiment Parameters}
\begin{table}[H]
\centering
\caption{Cost weights used in BiCompoDiff.}
\label{tab:cost_weights}
\begin{tabular}{l r}
\toprule
\textbf{Cost Weight} & \textbf{Value} \\
\midrule
Smoothness (pick \& place) $w_{\text{smooth}}^{\text{(PK, PL)}}$ & 0.005 \\
Smoothness (handover) $w_{\text{smooth}}^{\text{HO}}$ & 0.01 \\
Grasp collision $w_{\text{coll}}^{\text{(PK, PL)}}$ & 0.025 \\
Handover region $w_{\text{region}}^{\text{HO}}$ & 0.1 \\
Handover orientation $w_{\text{orient}}^{\text{HO}}$& 0.3 \\
Handover collision $w_{\text{coll}}^{\text{HO}}$ & 0.0001 \\
Regrasp $w_{\text{regrasp}}^{\text{(PK, PL)}}$ & 0.001 \\
\bottomrule
\end{tabular}
\end{table}

\subsection{Ablation studies}

\begin{figure}[!ht]
  \centering
 \includegraphics[width={0.45\linewidth}]{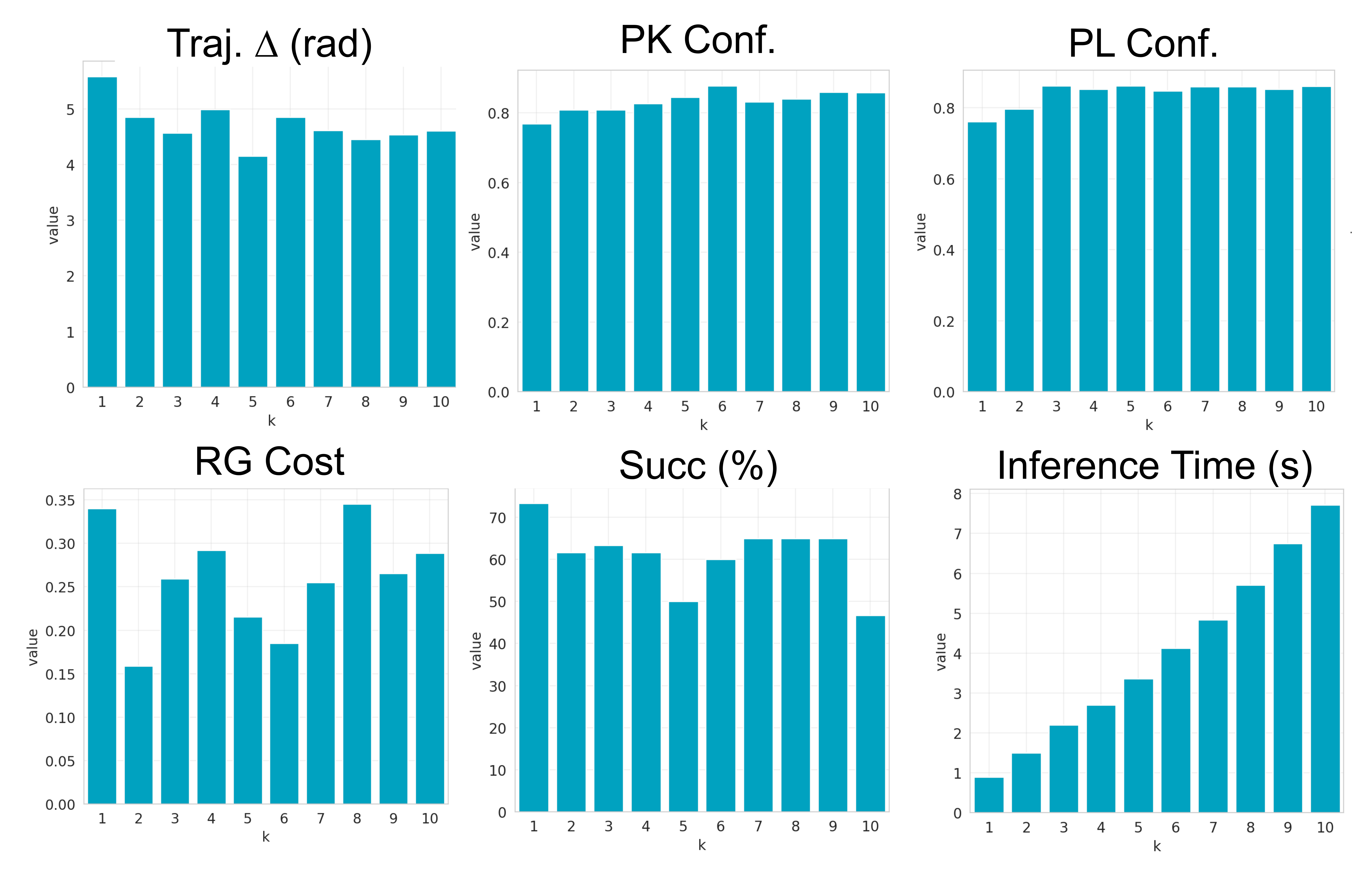}
  \caption{Performance by $k$ (Full mode + MCMC, all tasks).}
  \label{fig:k-ablation}
\end{figure}
\textbf{Inference time difference} in Table~\ref{tab:opt_modes_and_mcmc}. NoEBM: \textbf{0.29s}; BiCompoDiff-Full: 
\textbf{2.20s} (NoMCMC: 0.63s).

\end{document}